\pdfoutput=1

\documentclass[11pt]{article}

\usepackage{eeml2021}

\usepackage{xurl}

\usepackage{graphicx}
\usepackage{caption}
\usepackage{subcaption}

\usepackage{times}
\usepackage{latexsym}

\usepackage[T1]{fontenc}

\usepackage[utf8]{inputenc}

\usepackage{microtype}

\title{
Semantic Segmentation Alternative Technique:\\
Segmentation Domain Generation}

\author{
  Ana-Cristina Rogoz\thanks{\hspace{1em} All authors have equal contribution.}\\
  University of Bucharest\\
  \texttt{anacrogoz@gmail.com} 
     \And  
    Radu Muntean\footnotemark[1] \\
   ETH Z\"urich \\
   \texttt{radu.muntean.2@gmail.com}
   \AND
  Ștefan Cobeli\footnotemark[1] \\
  \texttt{stefancobeli@gmail.com}
  \\}

\begin{document}
\maketitle

\begin{abstract}
    Detecting objects of interest in images was always a compelling task to automate.
    In recent years this task was more and more explored using deep learning techniques, mostly using region-based convolutional networks.
    In this project we propose an alternative semantic segmentation technique making use of Generative Adversarial Networks.
    We consider semantic segmentation to be a domain transfer problem.
    Thus, we train a feed forward network (FFNN) to receive as input a seed real image and generate as output its segmentation mask. 
\end{abstract}

\section{Introduction}


    The problem of automatic detection and segmentation of objects in images can be encountered in many domains, from sports to medicine or physics \cite{ruhle2021workflow}. 
    Traditionally, one or more human experts would be required to perform live detection and give their expertise on the objects of interest location.
    Thus, automating this process was and still is considered of great value.
    
\begin{figure}
\centering
\begin{subfigure}{.21\textwidth}
  \centering
  \includegraphics[width=.9\linewidth]{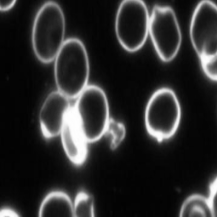}
  \caption{Source image}
  \label{fig:sub1}
\end{subfigure}%
\begin{subfigure}{.21\textwidth}
  \centering
  \includegraphics[width=.9\linewidth]{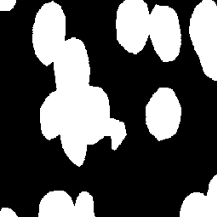}
  \caption{Segmentation mask}
  \label{fig:sub2}
\end{subfigure}
\caption{
Problem description: given a dataset labeled for a segmentation task, the goal is to train a \texttt{GAN} to be able to transform (a) the source image into its corresponding (b) segmentation mask.
}
\label{fig:problem_formulation}
\end{figure}

    In this project we propose a general method for automating the process of segmentation of desired objects in an image, as displayed in \autoref{fig:problem_formulation}.
    Our method is based on Generative Adversarial Networks (\texttt{GANs}) \citep{goodfellow2014generative}.

In recent years, the most successful methods for object segmentation were based on region proposal techniques such as \texttt{R-CNN} or \texttt{Fast R-CNN} \citep{girshick2015region, girshick2015fast}. 
Even though such techniques achieve high performance on many benchmark datasets, they reveal two specific drawbacks. 
First, they tend to run slower than a simple FFNN, even after their training phase. 
Second, as most deep learning techniques, they need large training datasets.

We propose to treat the segmentation problem as a domain transfer task.
The source domain is represented by the real images to be segmented and the target domain is represented by the segmentation masks.
By doing so, we can adapt established techniques for image translation such as Conditional Generative Adversarial Networks (\texttt{CGAN}) and  
Cycle-Consistent Generative Adversarial Networks (\texttt{CycleGAN})
 \cite{isola2017image, zhu2017unpaired} 
 for segmentation masks generation.

By using \texttt{GANs} for learning how to generate a segmentation mask we seek to address both aforementioned issues of the region proposal techniques.
First, since after the training phase we make use only of the \texttt{GAN}'s generator, the segmentation generation time is equal to that of a forward pass through a FFNN.
Second, both the \texttt{CGAN} and the \texttt{CycleGAN} revealed favorable results on relatively small datasets, with under $1,000$ samples, for example on the \textit{CMP Facades} dataset, as shown in \cite{isola2017image, zhu2017unpaired}.

\section{Datasets}

We seek to validate our 
methodology by testing it on two datasets.
The first dataset \texttt{Particles} contains $40$ images of particles acquired using electron microscopy \citep{ruhle2021workflow}.
Each image has an associated segmentation mask corresponding to the location of the particles.
The second dataset \texttt{Bacteria} contains $366$ images of the \textit{Spirochaeta} bacteria acquired using darkfield microscopy\footnote{\url{https://www.kaggle.com/longnguyen2306/bacteria-detection-with-darkfield-microscopy}}.
The location of the bacteria is marked with a corresponding segmentation mask.

\section{Experiments}
\label{sec:Experiments}
We led experiments on both datasets using \texttt{CGANs} and \texttt{CycleGANs}. 
For all the experiments we splitted our datasets in train and test sets. 
For \texttt{Particles}, $35$ samples for training and $5$ samples for testing.
For \texttt{Bacteria}, $320$ samples for training and $46$ samples for testing.

\subsection{Conditional GAN}

\begin{figure}[ht]
    \begin{tabular}{llll}
        \includegraphics[width=.275\linewidth]{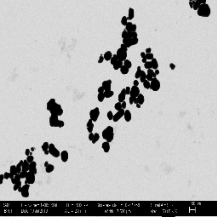} & \includegraphics[width=.275\linewidth]{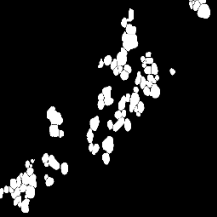} & \includegraphics[width=.275\linewidth]{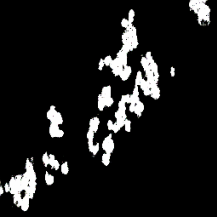}\\
        \includegraphics[width=.275\linewidth]{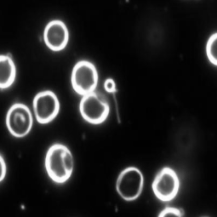} & \includegraphics[width=.275\linewidth]{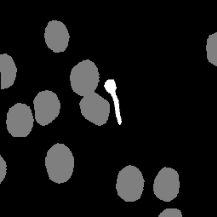} & \includegraphics[width=.275\linewidth]{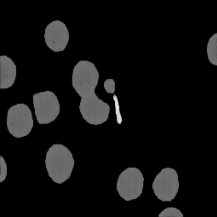}\\
    \end{tabular}
    \caption{Segmentation masks generated using \texttt{CGANs}.
    On each row we can see, from left to right: the real image, the ground truth segmentation mask and the generated segmentation mask.
    On the first and second row we can see examples from the \texttt{Particles} and \texttt{Bacteria} datasets respectively.
    }
    \label{fig:CGAN_generation_samples}
\end{figure}

We have tested a \texttt{CGAN} composed of a \texttt{U-Net} \cite{DBLP:journals/corr/RonnebergerFB15} generator and a Convolutional Neural Network (\texttt{2D-Conv})  discriminator. 
The model was trained for $100$ epochs. 
The last few epochs revealed stabilized values for the loss functions for both generator and discriminator.
Thus, $100$ epochs were enough for reaching a \textit{Nash Equilibrium} between the generator and the discriminator on the analyzed datasets.
We can see the \texttt{CGAN} results in \autoref{fig:CGAN_generation_samples}.

\subsection{Cycle GAN}

We used the same \texttt{U-Net} generator and \texttt{2D-Conv} discriminator when using the \texttt{CycleGAN}, as for the above mentioned experiments for the \texttt{CGAN}. 

The image translation problem setting has two main changes 
when using a \texttt{CycleGAN} compared to using a \texttt{CGAN}. 
First, a \texttt{CycleGAN} does not need \textit{(image, mask)} pairs for the translation.
It is enough to have a number of samples from a source \texttt{Domain A} and a number of samples from a target \texttt{Domain B}, whereas a \texttt{CGAN} requires perfect pairs between the two domains. 
Second, for a \texttt{CycleGAN} we have to train four discriminators and one generator. 
Thus the training procedure is more laborious.

\begin{figure}[ht]
    \begin{tabular}{llll}
        \includegraphics[width=.275\linewidth]{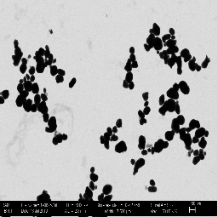} & \includegraphics[width=.275\linewidth]{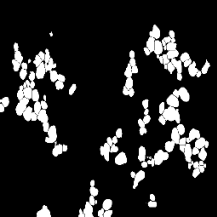} & \includegraphics[width=.275\linewidth]{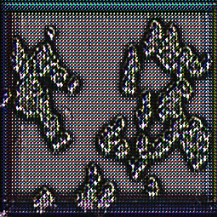}\\
        \includegraphics[width=.275\linewidth]{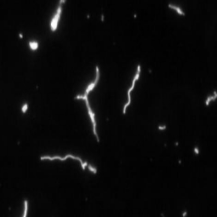} & \includegraphics[width=.275\linewidth]{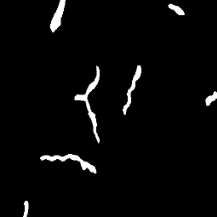} & \includegraphics[width=.275\linewidth]{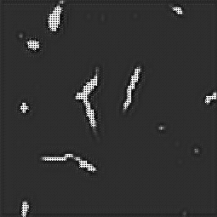}\\
    \end{tabular}
    \caption{Samples of segmentation masks generated using \texttt{CycleGANs}.
    On each row we can see, from left to right: the real image, the ground truth segmentation mask and the generated segmentation mask.
    On the first and second row we can see examples from the \texttt{Particles} and \texttt{Bacteria} datasets respectively.
    }
    \label{fig:CycleGAN_generation_samples}
\end{figure}

We have observed that the results of a trained \texttt{CycleGAN}  are inferior to those obtained by a trained \texttt{CGAN} as we can see in \autoref{fig:CycleGAN_generation_samples}. 
Since a \texttt{CycleGAN} does not take into account the \textit{(image, mask)} pairs, it is to be expected that the generations should be affected in quality. 

\section{Conclusions and Future Work}
We have shown in our preliminary experiments that indeed the segmentation problem can be treated as a domain transfer task and we proved our concept by using two types of generative networks, \texttt{CGANs} and \texttt{CycleGANs} on the \texttt{Particles} and \texttt{Bacteria} datasets.   

In our future experiments we aim to compare established segmentation techniques against our method and compare them on multiple metrics, such as generator's veracity when using a small training set, as well as segmentation generation speed.

Moreover, we wish to investigate the feasibility of the reverse problem. 
Mainly, the plausibility of generating true images from segmentation masks.
There are multiple domains where the possibility of generating real images from categorical masks would represent a real asset.


\bibliography{eeml2021}
\bibliographystyle{acl_natbib}

\end{document}